\documentclass[10pt,twocolumn,letterpaper]{article}
\usepackage{titling}        

\usepackage[pagenumbers]{cvpr} 

\usepackage{graphicx}
\usepackage{amsmath}
\usepackage{amssymb}
\usepackage{booktabs}
\usepackage{bm}
\usepackage{multirow}
\usepackage{makecell}
\usepackage{array}
\usepackage{wrapfig}
\usepackage[dvipsnames]{xcolor}
\usepackage{colortbl}
\usepackage{graphicx, caption}
\usepackage[ruled,vlined]{algorithm2e}
\usepackage{xspace}
\usepackage{pifont}
\usepackage[accsupp]{axessibility}  

\usepackage[pagebackref,breaklinks,colorlinks]{hyperref}

\usepackage[capitalize]{cleveref}
\crefname{section}{Sec.}{Secs.}
\Crefname{section}{Section}{Sections}
\Crefname{table}{Table}{Tables}
\crefname{table}{Tab.}{Tabs.}

\definecolor{F7E0D5}{RGB}{245,240,255}
\colorlet{Light}{White!0!F7E0D5}
\newcommand{\CC}{\cellcolor{Light}}
\def\sota{\textit{state-of-the-art }}
\newtheorem{definition}{Definition}
\newcommand{\na}{\textcolor{gray}{-}}
\newcommand{\cmark}{\ding{51}\xspace}
\newcommand{\xmark}{\ding{55}\xspace}
\def \pzo {\phantom{0}}
\makeatletter
\def\@fnsymbol#1{\ensuremath{\ifcase#1\or \dagger\or \ddagger\or
   \mathsection\or \mathparagraph\or \|\or **\or \dagger\dagger
   \or \ddagger\ddagger \else\@ctrerr\fi}}
\makeatother

\def\cvprPaperID{} 
\def\confName{CVPR}
\def\confYear{2022}

\begin{document}

\title{Automated Progressive Learning for Efficient Training of Vision Transformers}

\author{%
Changlin Li\textsuperscript{1,2,3} \quad
Bohan Zhuang\textsuperscript{3}\thanks{Corresponding author.} \quad
Guangrun Wang\textsuperscript{4} \quad
Xiaodan Liang\textsuperscript{5} \quad
Xiaojun Chang\textsuperscript{2} \quad
Yi Yang\textsuperscript{6}\\
{\normalsize%
\textsuperscript{1}Baidu Research \quad
\textsuperscript{2}ReLER, AAII, University of Technology Sydney}\\
{\normalsize%
\textsuperscript{3}Monash University \quad
\textsuperscript{4}University of Oxford \quad
\textsuperscript{5}Sun Yat-sen University \quad
\textsuperscript{6}Zhejiang University}\\
{\tt\small%
changlinli.ai@gmail.com, bohan.zhuang@monash.edu, wanggrun@gmail.com,}\\
{\tt\small%
xdliang328@gmail.com, xiaojun.chang@uts.edu.au, yangyics@zju.edu.cn}
}
\maketitle

\begin{abstract}
Recent advances in vision Transformers (ViTs) have come with a voracious appetite for computing power,
highlighting the urgent need to develop efficient training methods for ViTs. Progressive learning, a training scheme where the model capacity grows progressively during training, has started showing its ability in efficient training. In this paper, we take a practical step towards efficient training of ViTs by customizing and automating progressive learning. First, we develop a strong manual baseline for progressive learning of ViTs, by introducing momentum growth (MoGrow) to bridge the gap brought by model growth. Then, we propose automated progressive learning (AutoProg), an efficient training scheme that aims to achieve lossless acceleration by automatically increasing the training overload on-the-fly; this is achieved by adaptively deciding whether, where and how much should the model grow during progressive learning. Specifically, we first relax the optimization of the growth schedule to sub-network architecture optimization problem, then propose one-shot estimation of the sub-network performance via an elastic supernet. The searching overhead is reduced to minimal by recycling the parameters of the supernet. Extensive experiments of efficient training on ImageNet with two representative ViT models, DeiT and VOLO, demonstrate that AutoProg can accelerate ViTs training by up to 85.1\% with no performance drop.\footnote{Code: \url{https://github.com/changlin31/AutoProg}.}
\end{abstract}

\vspace{-1em}
\section{Introduction}
\label{sec:intro}
With powerful high model capacity and large amounts of data, Transformers have dramatically improved the performance on many tasks in computer vision (CV) \cite{touvron2020deit, liu2021swin}.
The pioneering ViT model~\cite{dosovitskiy2021image}, scales the model size to 1,021 billion FLOPs, 250$\times$ larger than ResNet-50~\cite{he2016deep}.
Through pre-training on the large-scale JFT-3B dataset \cite{zhai2021scaling}, the recently proposed ViT model, CoAtNet~\cite{dai2021coatnet}, reached \sota performance, with about 8$\times$ training cost of the original ViT.
The rapid growth in training scale inevitably leads to higher computation cost and carbon emissions. As shown in~\cref{tab:cost_of_vision_models}, recent breakthroughs of vision Transformers have come with a voracious appetite for computing power, resulting in considerable growth of environmental costs. Thus, it becomes extremely important to make ViTs training tenable in computation and energy consumption.

\begin{figure}
    \centering
    \includegraphics[width=.9\linewidth]{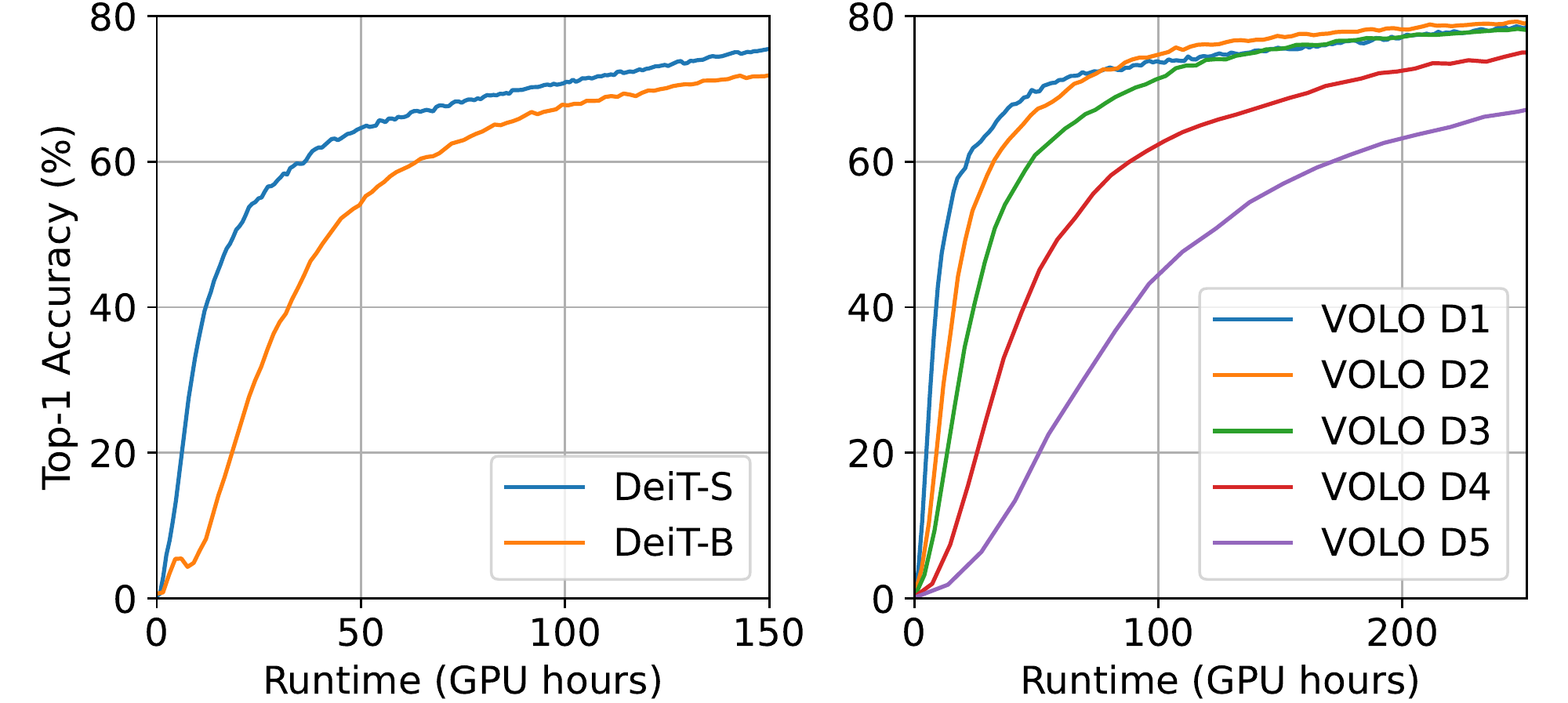}
    \caption{Accuracy of ViTs (DeiT~\cite{touvron2020deit} and VOLO~\cite{Yuan2021VOLOVO}) during training.
    Smaller ViTs converge faster in terms of runtime\protect\footnotemark.
    Models in the legend are sorted in increasing order of model size.}
    \label{fig:motivation1}
    \vspace{-2pt}
\end{figure}
\footnotetext{We refer runtime to the total GPU hours used in forward and backward pass of the model during training.}

\begin{table}[t]
\setlength{\tabcolsep}{3pt}
    \centering
    \scalebox{0.8}{
    \begin{tabular}{l|c|c}
        \toprule
         Model & CO$_2$e (lbs)\protect\footnotemark & ImageNet Acc. (\%) \\
         \midrule
         ResNet-50~\cite{he2016deep,dosovitskiy2021image} & 267 & 77.54\\
         \textcolor{gray}{BERT$_{base}$}~\cite{devlin2019bert} & \textcolor{gray}{1,438} & \textcolor{gray}{-} \\
         \textcolor{gray}{Avg person per year}~\cite{strubell2019energy} & \textcolor{gray}{11,023}& \textcolor{gray}{-} \\
         ViT-H/14~\cite{dosovitskiy2021image} & 22,793 & 88.55\\
         CoAtNet~\cite{dai2021coatnet} & 183,256 & 90.88\\
         \bottomrule
    \end{tabular}}
    \vspace{-5pt}
    \caption{The growth in training scale of vision models results in considerable growth of environmental costs. The CO$_2$e of human life and a language model, BERT~\cite{devlin2019bert} are also included for comparison. The results of ResNet-50, ViT-H/14 are from~\cite{dosovitskiy2021image}, and trained on JFT-300M~\cite{Sun2017RevisitingUE}. CoAtNet is trained on JFT-3B~\cite{zhai2021scaling}.} 
    \label{tab:cost_of_vision_models}
    \vspace{-17pt}
\end{table}
\footnotetext{CO$_2$ equivalent emissions (CO$_2$e) are calculated following~\cite{Patterson2021CarbonEA}, using U.S. average energy mix, \ie, 0.429 kg of CO$_2$e/KWh.}

In mainstream deep learning training schemes, all the network parameters participate in every training iteration. However, we empirically found that training only a small part of the parameters yields comparable performance in early training stages of ViTs. As shown in \cref{fig:motivation1},
smaller ViTs converge much faster in terms of runtime (though they would be eventually surpassed given enough training time). The above observation motivates us to rethink the efficiency bottlenecks of training ViTs:
does every parameter, every input element need to participate in all the training steps?

Here, we make the \textit{Growing Ticket Hypothesis} of ViTs:
the performance of a large ViT model, can be reached by first training its sub-network, then the full network after properly growing, with the same total training iterations.
This hypothesis generalizes the \textit{lottery ticket hypothesis}~\cite{frankle2019lottery} by adding a finetuning procedure at the full model size,
changing its scenario from \textit{efficient inference} to \textit{efficient training}. By iteratively applying this hypothesis to the sub-network, we have the progressive learning scheme.

Recently, progressive learning has started showing its capability in accelerating model training.
In the field of NLP, progressive learning can reduce half of BERT pre-training time~\cite{Gong2019EfficientTO}.
Progressive learning also shows the ability to reduce the training cost for convolutional neural networks (CNNs)~\cite{Tan2021EfficientNetV2SM}.
However, these algorithms differ substantially from each other, and their generalization ability among architectures is not well studied.
For instance, we empirically observed that progressive stacking~\cite{Gong2019EfficientTO} could result in significant performance drop (about 1\%) on ViTs.

To this end, we take a practical step towards sustainable deep learning by generalizing and automating progressive learning on ViTs. To the best of our knowledge, we are among the pioneering works to tackle the efficiency bottlenecks of training ViT models. We formulate progressive learning with its two components, \textit{growth operator} and \textit{growth schedule}, and study each component separately by controlling the other.

First, we present a strong manual baseline for progressive learning of ViTs by developing the growth operator. To ablate the optimization of the growth operator, we introduce a uniform linear growth schedule in two dimensions of ViTs, \ie, number of patches and network depth. To bridge the gap brought by model growth, we propose \textit{momentum growth} (\textbf{MoGrow}) operator with an effective momentum update scheme.
Moreover, we present a novel \textit{automated progressive learning} (\textbf{AutoProg}) algorithm that achieves lossless training acceleration by automatically adjusting the training overload.
Specifically, we first relax the optimization of the growth schedule to sub-network architecture optimization problem. Then, we propose one-shot estimation of sub-network performance via training an \textit{elastic supernet}. The searching overhead is reduced to minimal by recycling the parameters of the supernet.

The proposed AutoProg achieves remarkable training acceleration for ViTs on ImageNet. Without manually tuning, it consistently speeds up different ViTs training by more than 40\%, on disparate variants of DeiT and VOLO, including DeiT-tiny and VOLO-D2 with 72.2\% and 85.2\% ImageNet accuracy, respectively. Remarkably, it accelerates VOLO-D1 \cite{Yuan2021VOLOVO} training by up to 85.1\% with no performance drop. 
While significantly saving training time, AutoProg achieves competitive results when testing on larger input sizes and transferring to other datasets compared to the regular training scheme.

Overall, our contributions are as follows:
\begin{itemize}
\vspace{-4pt}
\itemsep -0.125cm
    \item 
    We develop a strong manual baseline for progressive learning of ViTs, by introducing MoGrow, a momentum growth strategy to  bridge  the  gap  brought  by  model  growing.
    \item We propose automated progressive learning (AutoProg), an efficient training scheme that aims to achieve lossless acceleration by automatically adjusting the growth schedule on-the-fly.
    \item Our AutoProg achieves remarkable training acceleration (up to 85.1\%) for ViTs on ImageNet, performing favourably against the original training scheme. 
\end{itemize}

\vspace{-1em}
\section{Related Work}
\label{sec:related_work}
\vspace{-0.5em}
\noindent\textbf{Progressive Learning.}
Early works on progressive learning~\cite{Fahlman1989TheCL,Lengell1996TrainingML,Hinton2006AFL,Bengio2006GreedyLT,Simonyan2015VeryDC,Smith2016GradualDO,Karras2018ProgressiveGO,Wang2017DeepGL} mainly focus on circumventing the training difficulty of deep networks.
Recently, as training costs of modern deep models are becoming formidably expensive, progressive learning starts to reveal its ability in \textit{efficient training}.
Net2Net~\cite{Chen2016Net2NetAL} and Network Morphism \cite{Wei2016NetworkM,Wei2017ModularizedMO,Wei2021ModularizedMO} studied how to accelerate large model training by properly initializing from a smaller model.
In the field of NLP, many recent works accelerate BERT pre-training by progressively stacking layers~\cite{Gong2019EfficientTO,Li2020ShallowtoDeepTF,Yang2020ProgressivelyS2}, dropping layers~\cite{zhang2020accelerating} or growing in multiple network dimensions~\cite{Gu2021OnTT}. Similar frameworks have also been proposed for efficient training of other models~\cite{You2020L2GCNLA,Wang2021StackRecET}.
As these algorithms remain hand-designed and could perform poorly when transferred to other networks, we propose to automate the design process of progressive learning schemes.

\noindent\textbf{Automated Machine Learning.}
Automated Machine Learning (AutoML) aims to automate the design of model structures and learning methods
from many aspects, including Neural Architecture Search (NAS)~\cite{zoph2016neural,baker2016designing,Tan2018MnasNetPN,liu2018progressive}, Hyper-parameter Optimization (HPO)~\cite{Bergstra2011AlgorithmsFH,Bergstra2012RandomSF}, AutoAugment~\cite{Cubuk2019AutoAugmentLA,Cubuk2020RandaugmentPA}, AutoLoss~\cite{Wu2018LearningTT,Xu2019AutoLossLD,li2020autosegloss}, \etc By relaxing the bi-level optimization problem in AutoML, there emerges many \textit{efficient AutoML algorithms} such as weight-sharing NAS~\cite{Liu2018DARTSDA,Cai2018ProxylessNASDN,brock2017smash,pham2018enas,guo2020single,li2019blockwisely,peng2021pi}, differentiable AutoAug~\cite{Li2020DADADA}, \etc These methods share network parameters in a jointly optimized \textit{supernet} for different candidate architectures or learning methods,
then rate each of these candidates according to its parameters inherited from the supernet. 

Attempts have also been made on \textit{automating progressive learning}.
AutoGrow~\cite{Wen2020AutoGrowAL} proposes to use a \textit{manually-tuned} progressive learning scheme to search for the optimal network depth, which is essentially a NAS method.
LipGrow~\cite{Dong2020TowardsAR} could be the closest one related to our work, which adaptively
decide the proper time to double the depth of CNNs on small-scale datasets, based on Lipschitz constraints. Unfortunately, LipGrow can not generalize easily to ViTs, as self-attention is not Lipschitz continuous~\cite{Kim2021TheLC}. In contrast, we solve the automated progressive learning problem from a traditional AutoML perspective, and fully automate the growing schedule by adaptively deciding \textit{whether}, \textit{where} and \textit{how much} to grow. Besides, our study is conducted directly on large-scale ImageNet dataset, in accord with practical application of efficient training.

\noindent\textbf{Elastic Networks.}
Elastic Networks, or anytime neural networks, are supernets executable with their sub-networks in various sizes, permitting instant and adaptive accuracy-efficiency trade-offs at runtime. Earlier works on Elastic Networks can be divided into \textit{Networks with elastic depth} \cite{larsson2016fractalnet,Huang2018MultiScaleDN,hu2019learning}, and \textit{networks with elastic width}~\cite{Lee2018AnytimeNP,yu2019slimmable,Yu2019UniversallySN}.
The success of elastic networks is followed by their two main applications, \textit{one-shot single-stage NAS}~ \cite{yu2019autoslim,Cai2020Once_for_All,Yu2020BigNASSU,chen2021autoformer} and \textit{dynamic inference}~\cite{Huang2018MultiScaleDN,Li2019ImprovedTF,li2021dynamic,wang2021not,li2021ds}, where emerges numerous elastic networks on \textit{multiple dimensions} (\eg, kernel size of CNNs~\cite{Cai2020Once_for_All,Yu2020BigNASSU}, head numbers~\cite{chen2021autoformer,hou2020dynabert} and patch size~\cite{wang2021not} of Transformers). From a new perspective, we present an elastic Transformer serving as a sub-network performance estimator during growth for automated progressive learning.%

\vspace{-8pt}
\section{Progressive Learning of Vision Transformers}
\vspace{-0.5em}

\begin{figure*}[t]
    \centering
    \includegraphics[width=.95\linewidth]{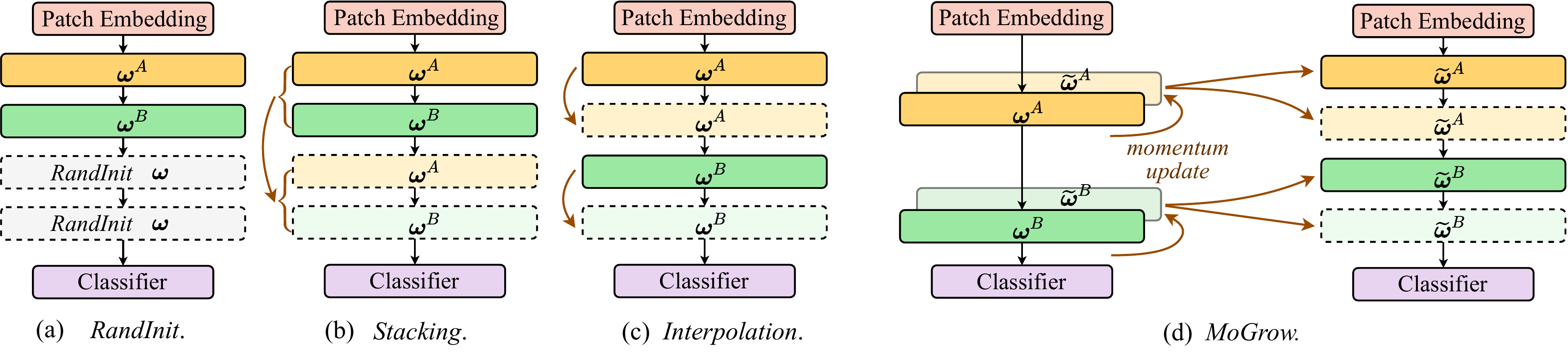}
    \vspace{-.5em}
    \caption{Variants of the growth operator $\bm\zeta$. $\bm\omega^\textit{A}$ and $\bm\omega^\textit{B}$ denote the parameters of two Transformer blocks in the original small network $\bm\psi_s$. (a) \textit{RandInit} randomly initializes newly added layers; (b) \textit{Stacking} duplicates the original layers and directly stacks the duplicated ones on top of them; (c) \textit{Interpolation} interpolate new layers of $\bm\psi_\ell$ in between original ones and copy the weights from their nearest neighbor in $\bm\psi_s$. (d) Our proposed \textit{MoGrow} is build upon \textit{Interpolation}, by coping parameters $\widetilde{\bm\omega}$ from the momentum updated ensemble of $\bm\psi_s$.}
    \label{fig:growth_operators}
    \vspace{-1.5em}
\end{figure*}

In this section, we aim to develop a strong manual baseline for progressive learning of ViTs. 
We start by formulating progressive learning with its two main factors, \textit{growth schedule} and \textit{growth operator} in Sec.~\ref{sec:Prog}. Then, we present the growth space that we use in Sec.~\ref{sec:GS}. Finally, we explore the most suitable growth operator of ViTs in Sec.~\ref{sec:GO}.

\noindent\textbf{Notations.} We denote scalars, tensors 
and sets (or sequences) using lowercase, bold lowercase and uppercase letters (\eg, $n$, $\bm x$ and $\Psi$). For simplicity, we use $\{\bm x_n\}$ to denote the set $\{\bm x_n\}^{|n|}_{n=1}$ with cardinality $|n|$, similarly for a sequence $\left(\bm x_n\right)^{|n|}_{n=1}$. Please refer to \cref{tab:notations}
for a vis-to-vis explanation of the notations we used.
\vspace{-0.3em}
\subsection{Progressive Learning} \label{sec:Prog}
\vspace{-0.3em}

Progressive learning gradually increases the training overload by growing among its sub-networks following
a \textit{growth schedule} $\Psi$, which can be denoted by a sequence of sub-networks with increasing sizes for all the training epochs~$t$.
In practice, to ensure the network is sufficiently optimized after each growth, it is a common practice~\cite{Yang2020ProgressivelyS2,Gu2021OnTT,Tan2021EfficientNetV2SM} to divide the whole training process into $|k|$ equispaced stages with $\tau = |t|/|k|$ epochs in each stage. Thus, the growth schedule can be denoted as ${\Psi = \Big(\bm{\psi}_k\Big)_{k=1}^{|k|}}$; the final one is always the complete model.
Note that stages with different lengths can be achieved by using the same $\bm\psi$ in different numbers of consecutive stages, \eg, ${\Psi = (\bm\psi_a, \bm\psi_b, \bm\psi_b)}$, where $\bm\psi_a, \bm\psi_b$ are two different sub-networks.

When growing a sub-network to a larger one, the parameters of the larger sub-network are initialized by a \textit{growth operator} $\bm\zeta$, which is a reparameterization function that maps the weights $\bm\omega_s$ of a smaller network to $\bm\omega_\ell$ of a larger one by ${\bm\omega_\ell=\bm\zeta(\bm\omega_s)}$.
The whole progress of progressive learning is summarized in \cref{alg:prog}.
\setlength{\textfloatsep}{2pt}
\begin{table}[t]
    \centering
    \footnotesize
    \setlength{\tabcolsep}{4pt}
    \begin{tabular}{l|c|l}
    \toprule
    Notation        & Type & Description  \\
    \midrule
    $t$, $|t|$      & scalar    & training epoch, total training epochs \\
    $k$, $|k|$      & scalar    & training stage, total training stages \\
    $\tau$          & scalar    & epochs per stage \\
    $\Psi$          & sequence  & growth schedule \\
    $\bm\zeta$      & function   & growth operator\\
    $\bm\psi$       &   network        & sub-network \\
    $\Phi$          &  network         & supernet \\
    $\bm\omega$, $|\bm\omega|$     &  parameter         & network parameters, number of parameters\\
    $\Omega$, $\Lambda$         &   set    & the whole growth space, partial growth space\\
    $*$, $\star$       &    notation       & optimal, relaxed optimal\\
    \bottomrule
    \end{tabular}
    \vspace{-5pt}
    \caption{Notations describing progressive learning and automated progressive learning.}
    \vspace{-8pt}
    \label{tab:notations}
\end{table}
\begin{algorithm}[t]
\footnotesize
\SetAlgoLined
\textbf{Input:}\\  
$\Psi$:~the~growth~schedule; 
$\bm\zeta$:~the~growth~operator;\\
$|t|$:~total~training~epochs;
$\tau$:~epochs per stage;\\
Randomly initialized parameters $\bm\omega$;\\
\For {$t \in [1,|t|]$}{
    \If{~~$t=N\tau,~~N\in\mathbb{N_{+}}$~~}{
    Switch to the sub-network of next stage~~$\bm\psi\leftarrow\Psi[t/\tau]$;\\
    Initialize parameters by growth operator~~$\bm\omega\leftarrow\bm\zeta(\bm\omega)$;}{}
    Train $\bm\psi(\bm\omega)$ over all the training data.
}
\caption{Progressive Learning}
\label{alg:prog}
\end{algorithm}

Let $\mathcal{L}$ be the target loss function,
and $\mathcal{T}$ be the total runtime; then progressive learning can be formulated as:
\begin{equation}\label{eq:prog}
\setlength{\abovedisplayskip}{3pt} 
\setlength{\belowdisplayskip}{3pt}
    \mathop{\min}_{\bm\omega, \Psi, \bm\zeta}\big\{\mathcal{L}(\bm\omega, \Psi, \bm\zeta), \mathcal{T}(\Psi)\big\},
\end{equation}
where $\bm\omega$ denotes the parameters of sampled sub-networks during the optimization.
Growth schedule $\Psi$ and growth operator $\bm\zeta$ have been explored for language Transformers~\cite{Gong2019EfficientTO,Gu2021OnTT}. However, ViTs differ considerably from their linguistic counterparts. The huge difference on task objective, data distribution and network architecture could lead to drastic difference in optimal $\Psi$ and $\bm\zeta$. In the following parts of this section, we mainly study the growth operator $\bm\zeta$ for ViTs by fixing $\Psi$ as a \textit{uniform linear schedule} in a \textit{growth space} $\Omega$, and leave automatic exploration of $\Psi$ to~\cref{sec:autoprog}.

\vspace{-0.3em}
\subsection{Growth Space in Vision Transformers} \label{sec:GS}
\vspace{-0.3em}

The model capacity of ViTs are controlled by many factors, such as number of patches, network depth, embedding dimensions, MLP ratio, \etc. In analogy to previous discoveries on fast compound model scaling~\cite{Dollr2021FastAA}, we empirically find that reducing network width (\eg, embedding dimensions) yields relatively smaller wall-time acceleration on modern GPUs when comparing at the same \textit{flops}. Thus, we mainly study \textit{number of patchs ($n^2$)} and \textit{network depth~($l$)}, leaving other dimensions for future works.

\noindent\textbf{Number of Patches.}
Given patch size $p\times p$, input size ${r\times r}$, the number of patches $n\times n$ is determined by $n^2 = r^2/p^2$. Thus, by fixing the patch size,
reducing \textit{number of patches} can be simply achieved by
down-sampling the input image. 
However, in ViTs, the size of positional encoding is related to~$n$. To overcome this limitation, we adaptively interpolate the positional encoding to match with~$n$.

\noindent\textbf{Network Depth.}
Network depth ($l$) is the number of Transformer blocks or its variants (\eg, Outlooker blocks~\cite{Yuan2021VOLOVO}). 

\noindent\textbf{Uniform Linear Growth Schedule.}
To ablate the optimization of growth operator $\bm\zeta$, we fix growth schedule $\Psi$ as a \textit{uniform linear growth schedule}.
To be specific, \textit{``uniform''} means that all the dimensions (\ie, $n$ and $l$) are scaled by the same ratio $s_t$ at the $t$-th epoch; \textit{``linear''} means that the ratio~$s$
grows linearly from $s_1$~to~$1$.
This manual schedule has only one hyper-parameter, the initial scaling ratio~$s_1$, which is set to $0.5$ by default. 
With this fixed $\Psi$, the optimization of progressive learning in \cref{eq:prog} is simplified to:
\begin{equation}\label{eq:zeta}
\setlength{\abovedisplayskip}{3pt} 
\setlength{\belowdisplayskip}{3pt}
    \mathop{\min}_{\bm\omega, \bm\zeta}\mathcal{L}(\bm\omega, \bm\zeta),
\end{equation}
which enables direct optimization of $\bm\zeta$ by comparing the final accuracy after training with different $\bm\zeta$.

\vspace{-0.3em}
\subsection{On the Growth of Vision Transformers} \label{sec:GO}
\vspace{-0.3em}
\begin{figure*}[t]
    \centering
    \includegraphics[width=.95\linewidth]{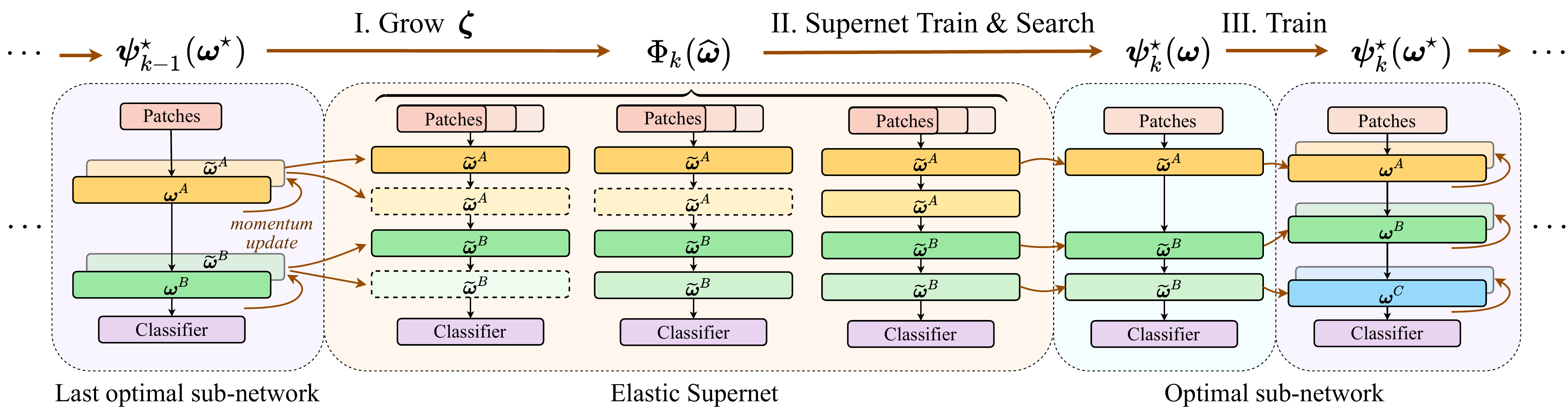}
    \vspace{-0.5em}
    \caption{Pipeline of the $k$-th stage of automated progressive learning. In the beginning of the stage, the last optimal sub-network $\bm\psi^\star_{k-1}$ first grows to the Elastic Supernet $\Phi_k$ by $\widehat{\bm\omega} = \bm\zeta(\bm\omega^\star)$; then, we search for the optimal sub-network $\bm\psi^\star_{k}$ after supernet training; finally, the sub-network is trained in the remaining epochs of this stage. The whole process of AutoProg is summarized in \cref{alg:autoprog}.}
    \label{fig:autoprog}
    \vspace{-1.5em}
\end{figure*}

\cref{fig:growth_operators}
(a)-(c) depict the main variants of the growth operator $\bm\zeta$ that we compare, which cover choices from a wide range of the previous works, including \textit{RandInit}~\cite{Simonyan2015VeryDC}, \textit{Stacking}~\cite{Gong2019EfficientTO} and \textit{Interpolation}~\cite{chang2018multi,Dong2020TowardsAR}. 
More formal definitions of these schemes can be found in the supplementary material.
Our empirical comparison (in~\cref{sec:exp_ablation}) shows Interpolation growth is the most suitable scheme for ViTs. 

Unfortunately, growing by Interpolation changes the original function of the network.
In practice, function perturbation brought by growth can result in significant performance drop, which is hardly recovered in subsequent training steps. 
Early works advocate for function-preserving growth operators~\cite{Chen2016Net2NetAL,Wei2016NetworkM}, which we denote by \textit{Identity}. However, we empirically found growing by Identity greatly harms the performance on ViTs (see~\cref{sec:exp_ablation}).
Differently, we propose a growth operator, named \textit{Momentum Growth (MoGrow)}, to bridge the gap brought by model growth.

\noindent\textbf{Momentum Growth (MoGrow).}
In recent years, a growing number of self-supervised~\cite{BYOL,guo2020bootstrap,He2020MomentumCF} and semi-supervised~\cite{laine2016temporal,tarvainen2017mean} methods learn knowledge from the historical ensemble of the network. Inspired by this, we propose to transfer knowledge from a \textit{momentum network} during growth. This momentum network has the same architecture with $\bm\psi_s$ and its parameters $\bm{\widetilde{\omega}}_s$ are updated with the online parameters $\bm\omega_s$ in every training step by:
\begin{equation}
\setlength{\abovedisplayskip}{3pt} 
\setlength{\belowdisplayskip}{3pt}
    \bm{\widetilde{\omega}}_s \leftarrow m\bm{\widetilde{\omega}}_s + (1-m)\bm\omega_s,
\end{equation}
where $m$ is a momentum coefficient set to $0.998$. As the the momentum network usually has better generalization ability and better performance
during training, loading parameters from the momentum network would help the model bypass the function perturbation gap. As shown in \cref{fig:growth_operators} (d),
MoGrow is proposed upon Interpolation growth by maintaining a momentum network, and directly copying from it when performing network growth. MoGrow operator $\bm\zeta_\textit{MoGrow}$ can be simply defined as:
\begin{equation}
\setlength{\abovedisplayskip}{3pt} 
\setlength{\belowdisplayskip}{3pt}
    \bm\zeta_\textit{MoGrow}(\bm\omega_s) = \bm\zeta_\textit{Interpolation}\big(\bm{\widetilde{\omega}}_s\big).
\end{equation}

\vspace{-0.7em}
\section{Automated Progressive Learning}
\label{sec:autoprog}
\vspace{-0.5em}
\setlength{\textfloatsep}{2pt}

\begin{algorithm}[t]
\footnotesize
\SetAlgoLined
\textbf{Input:}\\  
$\bm\zeta$:~the~growth~operator;\\
$|t|$:~total~training~epochs;
$\tau$:~epochs per stage;\\
Random initialize parameters $\bm\omega$;\\
\For {$t \in [1,|t|]$}{
    \If{~~$t=N\tau,~~N\in\mathbb{N_{+}}$~~}{
    Switch optimizers to \textit{Elastic Supernet}~~$\Phi$;\\
    Initialize supernet parameters~~$\widehat{\bm\omega}\leftarrow\bm\zeta(\bm\omega)$;\\}{}
    \If{~~$t=N\tau + 2,~~N\in\mathbb{N_{+}}$~~}{
    Search for the \textit{optimal sub-network}~$\bm\psi$ by \cref{eq:objective_3};\\
    Switch to the \textit{optimal sub-network}~~$\bm\psi\leftarrow\bm\psi^\star$;\\
    Inherit parameters from the supernet~~$\bm\omega\leftarrow\widehat{\bm\omega}$;}{}
    Train $\bm\psi(\bm\omega)$ or supernet $\Phi(\widehat{\bm\omega})$ over all the training data.
}
\caption{Automated Progressive Learning}
\label{alg:autoprog}
\end{algorithm}

In this section, we focus on optimizing the growth schedule $\Psi$ by fixing the growth operator as $\bm\zeta_\textit{MoGrow}$. We first formulate the multi-objective optimization problem of $\Psi$, then propose our solution, called \textit{AutoProg}, which is introduced in detail by its two estimation steps in \cref{sec:subnet} and \cref{sec:supernet}.%
\vspace{-0.4em}
\subsection{Problem Formulation}\label{sec:AutoProg}
\vspace{-0.4em}
The problem of designing growth schedule $\Psi$ for efficient training is a multi-objective optimization problem~\cite{deb2014multi}. By fixing $\bm\zeta$ in \cref{eq:prog} as our proposed $\bm\zeta_\textit{MoGrow}$, the objective of designing growth schedule $\Psi$ is: 
\begin{equation}\label{eq:psi}
\setlength{\abovedisplayskip}{3pt} 
\setlength{\belowdisplayskip}{2pt}
    \mathop{\min}_{\bm\omega, \Psi}\big\{\mathcal{L}(\bm\omega, \Psi), \mathcal{T}(\Psi)\big\}.
\end{equation}

Note that multi-objective optimization problem has a set of Pareto optimal~\cite{deb2014multi} solutions which can be approximated using customized weighted product, a common practice used in previous Auto-ML algorithms~\cite{Tan2018MnasNetPN,Tan2019EfficientNetRM}. In the scenario of progressive learning, the optimization objective can be defined as:
\begin{equation}\label{eq:psi_product}
\setlength{\abovedisplayskip}{3pt} 
\setlength{\belowdisplayskip}{3pt}
    \mathop{\min}_{\bm\omega, \Psi}\mathcal{L}(\bm\omega, \Psi)\cdot \mathcal{T}(\Psi)^\alpha,
\end{equation}
where $\alpha$ is a balancing factor dynamically chosen by balancing the scores for all the candidate sub-networks. 

\vspace{-7pt}
\subsection{Automated Progressive Learning by Optimizing Sub-Network Architectures}\label{sec:subnet}
\vspace{-0.3em}
Denoting $|\bm\psi|$ the number of candidate sub-networks, and $|k|$ the number of stages, the number of candidate growth schedule is thus $|\bm{\psi}|^{|k|}$. As optimization of \cref{eq:psi_product} contains optimization of network parameters $\bm\omega$, to get the final loss, a full $|t|$ epochs training with growth schedule $\Psi$ is required:
\begin{equation}
\label{eq:objective}
\setlength{\abovedisplayskip}{3pt} 
\setlength{\belowdisplayskip}{2pt}
\begin{aligned}
    \Psi^* = \mathop{\arg\min}_{\Psi} \mathcal{L}\big(\bm{\omega}^*(\Psi); \bm{x}\big) \cdot \mathcal{T}(\Psi) ^{\alpha}\\
    {\text{s.t.}}~~~~\bm{\omega}^*(\Psi) = \mathop{\arg\min}_{\bm{\omega}} \mathcal{L}(\Psi, {\bm{\omega}}; \bm{x}).
\end{aligned}
\end{equation}
Thus, performing an extensive search over the higher level factor $\Psi$ in this bi-level optimization problem has complexity $\mathcal{O}(|\bm{\psi}|^{|k|}\cdot|t|)$. Its expensive cost
deviates from the original intention of efficient training. 

To reduce the search cost, we relax the original objective of growth schedule search to progressively deciding \textit{whether}, \textit{where} and \textit{how much} should the model grow, by
searching the optimal sub-network architecture $\bm\psi^*_k$ in each stage $k$. Thus, the relaxed optimal growth schedule can be denoted as ${\Psi^\star = \Big(\bm{\psi}^*_k\Big)_{k=1}^{|k|}}$. 

In sparse training and efficient Auto-ML algorithms, it is a common practice to estimate future ranking of models
with current parameters and their gradients~\cite{evci2020rigging,tanaka2020pruning}, or with parameters after a single step of gradient descent update~\cite{Liu2018DARTSDA,Cai2018ProxylessNASDN,Li2020DADADA}. However, these methods are not suitable for progressive training, as the network function is drastically changed and is not stable after growth.
We empirically found that the network parameters adapt quickly after growth and are already stable after one epoch of training.
To make a good tradeoff between accuracy and training speed,
we estimate the performance of each sub-network $\bm{\psi}$ in each stage by their training loss after the first \textit{two} training epochs in this stage. 
Denoting $\bm{\omega}^\star$ the sub-network parameters obtained by two epochs of training, the optimal sub-network can be searched by:
\begin{equation}
\label{eq:objective_2}
\setlength{\abovedisplayskip}{3pt} 
\setlength{\belowdisplayskip}{3pt}
\begin{aligned}
    &\bm\psi_k^* = \mathop{\arg\min}_{\bm{\psi}_k\in\Lambda_k}\mathcal{L}\big(\bm{\omega}^\star(\bm{\psi}_k);\bm{x}\big) \cdot \mathcal{T}(\bm{\psi}_k)^\alpha,\\
    \text{where}~~~~&\Lambda_k = \Big\{\bm\psi\in\Omega~\Big|~|\bm\omega(\bm\psi)| \geq |\bm\omega(\bm\psi_k)|\Big\},
\end{aligned}
\end{equation}
where $\Lambda_k$ denotes the growth space of the $k$-th stage,
containing all the sub-networks that are larger than or equal to the last optimal sub-network in terms of the number of parameters $|\bm\omega|$.

Overall, by relaxing the original optimization problem in \cref{eq:objective} to \cref{eq:objective_2}, we only have to train each of the ${|\Lambda_k|}$ sub-networks for two epochs in each of the $|k|$ stages. Thus, the search complexity is reduced exponentially from $\mathcal{O}(|\bm{\psi}|^{|k|}\cdot|t|)$ to $\mathcal{O}(|\Lambda_k|\cdot|k|)$, where $|\Lambda_k|\leq|\bm{\psi}|$ and $|k|\leq|t|$.

\vspace{-0.3em}
\subsection{One-shot Estimation of Sub-Network Performance via Elastic Supernet}\label{sec:supernet}
\vspace{-0.3em}

Though we relax the optimization problem with significant search cost reduction,
obtaining $\bm\omega^\star$ in \cref{eq:objective_2} still takes ${2|\Lambda_k|}$ epochs for each stage, which will bring huge searching overhead to the progressive learning. 
The inefficiency of loss prediction is caused by the repeated training of sub-networks weight $\bm{\omega}$, with bi-level optimization being its nature. 
To circumvent this problem, 
we propose to share and jointly optimize sub-network parameters in $\Lambda_k$ via an \textit{Elastic Supernet with Interpolation}. 

\noindent\textbf{Elastic Supernet with Interpolation.}
An Elastic Supernet $\Phi({\widehat{\bm\omega}})$ is a \textit{weight-nesting} network parameterized by $\widehat{\bm\omega}$, and is able to execute with its sub-networks $\{\bm\psi\}$. Here, we give the formal definition of \textit{weight-nesting}:
\begin{definition}
\vspace{-5pt}
(weight-nesting) For any pair of sub-networks $\bm\psi_a(\bm\omega_a)$ and $\bm\psi_b(\bm\omega_b)$ in supernet $\Phi$, where $|\bm\omega_a| \leq |\bm\omega_b|$, if $\bm\omega_a \subseteq \bm\omega_b$ is always satisfied, then $\Phi$ is \textbf{weight-nesting}.
\vspace{-5pt}
\end{definition}
In previous works, a network with elastic depth is usually achieved by using the first layers to form sub-networks~\cite{Huang2018MultiScaleDN,Yu2020BigNASSU,chen2021autoformer}. However, using this scheme after growing by \textit{Interpolation} or \textit{MoGrow} will cause inconsistency between expected sub-networks after growth and sub-networks in $\Phi$.

To solve this issue, we present an \textit{Elastic Supernet with Interpolation}, with optionally activated layers interpolated in between always activated ones. 
As shown in \cref{fig:autoprog},
beginning from the smaller network in the last stage $\bm\psi^\star_{k-1}$, sub-networks in $\Phi$ are formed by inserting layers in between the original layers of $\bm\psi^\star_{k-1}$ (starting from the final layers), until reaching the largest sub-network in $\Lambda_k$. 

\noindent\textbf{Training and Searching via Elastic Supernet.} 
By nesting parameters of all the candidate sub-networks in the Elastic supernet $\Phi$, the optimization of $\bm\omega$ is disentangled from $\bm\psi$. Thus, \cref{eq:objective_2} is further relaxed to
\begin{equation}\label{eq:objective_3}
\setlength{\abovedisplayskip}{3pt} 
\setlength{\belowdisplayskip}{3pt}
\begin{aligned}
&\bm\psi_k^\star = \mathop{\arg\min}_{\bm\psi_k\in\Lambda_k} \mathcal{L}\big(\widehat{\bm\omega}^\star; \bm{x}\big) \cdot \mathcal{T}(\bm\psi_k) ^{\alpha}\\
{\text{s.t.}}~~~~&\widehat{\bm\omega}^{\star} = \mathop{\arg\min}^{~}_{\widehat{\bm\omega}}\mathbb{E}_{\bm\psi_k\in \Lambda_k}\big\{\mathcal{L}(\bm\psi_k, \widehat{\bm\omega}; \bm{x})\big\},
\end{aligned}
\end{equation}
where the optimal nested parameters $\widehat{\bm\omega}^\star$ can be obtained by one-shot training of $\Phi$ for two epochs. For efficiency, we train $\Phi$ by randomly sampling only one of its sub-networks in each step (following~\cite{chen2021autoformer}), instead of four in~\cite{Yu2019UniversallySN,yu2019autoslim,Yu2020BigNASSU}. 

After training all the candidate sub-networks in the Elastic Supernet $\Phi$ concurrently for two epochs, we have the adapted supernet parameters $\widehat{\bm\omega}^\star$ that can be used to estimate the real performance of the sub-networks (\ie performance when trained in isolation). As the sub-network grow space $\Lambda_k$ in each stage is relatively small,
we can directly perform traversal search in $\Lambda_k$, by testing its training loss with a small subset of the training data. We use fixed data augmentation to ensure fair comparison, following
\cite{Li2021BossNASEH}. Benefiting from parameter nesting and one-shot training of all the sub-networks in $\Lambda_k$, the search complexity is further reduced from $\mathcal{O}(|\Lambda_k|\cdot|k|)$ to $\mathcal{O}(|k|)$.

\noindent\textbf{Weight Recycling.}
Benefiting from synergy of different sub-networks, the supernet converges at a comparable~speed to training these sub-networks in isolation.
Similar phenomenon can be observed in network regularization~\cite{Srivastava2014DropoutAS,Huang2016DeepNW}, network augmentation~\cite{Cai2021NetworkAF}, and previous elastic models~\cite{yu2019slimmable,Yu2020BigNASSU,chen2021autoformer}. Motivated by this, the searched sub-network directly inherits its parameters in the supernet to continue training.
Benefiting from this \textit{weight recycling} scheme, AutoProg has \textit{no} extra searching epochs, since the supernet training epochs are parts of the whole training epochs. 
Moreover, as sampled sub-networks are faster than the full network, these supernet training epochs take less time than the original training epochs.
Thus, the searching cost is directly reduced from $\mathcal{O}(|k|)$ to \textbf{\textit{zero}}.

\begin{table}[t]
    \footnotesize
    \centering
    \setlength{\tabcolsep}{4pt}
    \begin{tabular}{l|l|c|c|c}
        \toprule
        Model  & \makecell[l]{Training\\scheme}  & \makecell{Speedup\\runtime}    & \makecell{Top-1\\(\%)} & \makecell{Top-1@288\\(\%)}\\
        \midrule
        \multicolumn{4}{l}{\textbf{\textit{100 epochs}}}\\
        \midrule
        \multirow{3}{*}{DeiT-S~\cite{touvron2020deit}}          & Original     & \na           & 74.1      & 74.6\\ %
        & Prog                                                     & \textcolor{gray}{+53.6\%} & 72.6      &  73.2\\ %
        & \CC AutoProg                                                 & \CC +40.7\%                   & \CC \textbf{74.4}      & \CC \textbf{74.9}\\ %
        \arrayrulecolor{lightgray}\hline\arrayrulecolor{black}
        \multirow{4}{*}{VOLO-D1~\cite{Yuan2021VOLOVO}}          & Original     & \na           & 82.6      &83.0\\ %
        & Prog                                                     & \textcolor{gray}{+60.9\%} & 81.7      &82.1\\ %
        & \CC AutoProg 0.5$\Omega$                                     & \CC +65.6\%                   & \CC \textbf{82.8}      &\CC \textbf{83.2}\\ %
        & \CC AutoProg 0.4$\Omega$                                     & \CC \textbf{+85.1\%}          & \CC 82.7      & \CC 83.1 \\ %
        \arrayrulecolor{lightgray}\hline\arrayrulecolor{black}
        \multirow{3}{*}{VOLO-D2~\cite{Yuan2021VOLOVO}}          & Original     & \na                       & 83.6      & 84.1\\ %
        & Prog                                                     & \textcolor{gray}{+54.4\%} & 82.9      & 83.3\\ %
        & \CC AutoProg                                                 & \CC +45.3\%                   &\CC  \textbf{83.8}      &\CC \textbf{84.2}\\ %
        \midrule
        \multicolumn{4}{l}{\textbf{\textit{300 epochs}}}\\
        \midrule
        \multirow{2}{*}{DeiT-Tiny~\cite{touvron2020deit}}       &  Original    & \na           & 72.2      & 72.9\\ %
        &\CC AutoProg                                                 & \CC    +51.2\%                       &\CC  \textbf{72.4}  &\CC \textbf{73.0}  \\ %
        \arrayrulecolor{lightgray}\hline\arrayrulecolor{black}
        \multirow{2}{*}{DeiT-S~\cite{touvron2020deit}}          &  Original    & \na           & 79.8      &  80.1\\ %
        & \CC AutoProg                                                 &\CC +42.0\%                   &\CC 79.8      & \CC 80.1 \\ %
        \arrayrulecolor{lightgray}\hline\arrayrulecolor{black}
        \multirow{2}{*}{VOLO-D1~\cite{Yuan2021VOLOVO}}          & Original     & \na           & 84.2      & 84.4 \\ %
        & \CC AutoProg                                                 & \CC +48.9\%                   &\CC \textbf{84.3}      &\CC \textbf{84.6}\\ %
        \arrayrulecolor{lightgray}\hline\arrayrulecolor{black}
        \multirow{2}{*}{VOLO-D2~\cite{Yuan2021VOLOVO}}          &Original      & \na           & 85.2      & 85.1\\
        &\CC AutoProg                                                 &\CC +42.7\%                   &\CC 85.2     & \CC \textbf{85.2}\\
        \bottomrule
    \end{tabular}%
    \caption{Main results of efficient training on ImageNet. Accelerations that cause accuracy drop are marked with \textcolor{gray}{gray}. Best results are marked with \textbf{Bold}; our method or default settings are highlighted in \colorbox{Light}{purple}. Top-1@288 denotes Top-1 Accuracy when directly testing on 288$\times$288 input size, \textit{without} finetuning. Please refer to the supplementary file for detailed FLOPs and runtime.}
    \label{tab:imagenet}
    \vspace{2pt}
\end{table}
\begin{figure*}[t]
    \vspace{-5pt}
    \centering
    \includegraphics[width=\linewidth]{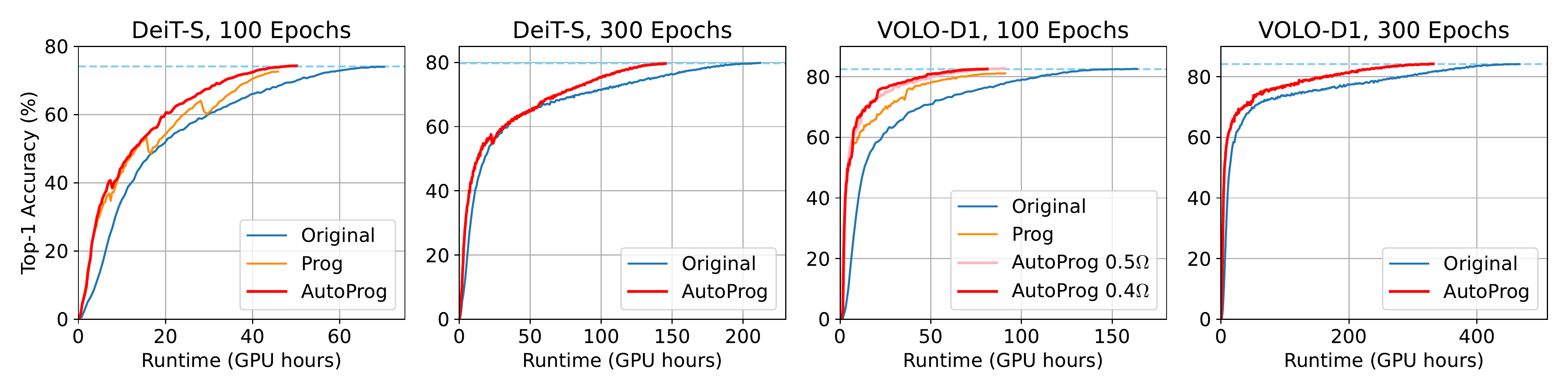}
    \vspace{-2em}
    \caption{Evaluation accuracy of DeiT-S and VOLO-D1 during training with different learning schemes. Curves are \textit{not} smoothed.}
    \vspace{-1em}
    \label{fig:learning_curve}
\end{figure*}
\vspace{-0.5em}
\section{Experiments}
\vspace{-0.5em}
\noindent\textbf{Datasets.}
We evaluate our method on a large scale image classification dataset, ImageNet-1K~\cite{deng2009imagenet} and two widely used classification datasets, CIFAR-10 and CIFAR-100~\cite{Krizhevsky09cifar}, for transfer learning. ImageNet contains 1.2M train set images and 50K val set images in 1,000 classes. We use all the training data for progressive learning and supernet training, and use a 50K randomly sampled subset to calculate training loss for sub-network search.

\noindent\textbf{Architectures.}
We use two representative ViT architectures, DeiT~\cite{touvron2020deit} and VOLO~\cite{Yuan2021VOLOVO} to evaluate the proposed AutoProg.
Specifically, DeiT~\cite{touvron2020deit} is a representative standard ViT model; 
VOLO~\cite{Yuan2021VOLOVO} is a hybrid architecture comprised of \textit{outlook attention} blocks and transformer blocks.

\noindent\textbf{Implementation Details.}
For both architectures, we use the original training hyper-parameters, data augmentation and regularization techniques of their corresponding prototypes~\cite{touvron2020deit,Yuan2021VOLOVO}. Our experiments are conducted on NVIDIA 3090 GPUs.
As the acceleration achieved by our method is orthogonal to the acceleration of mixed precision training~\cite{micikevicius2018mixed}, we use it in both original baseline training and our progressive learning.

\noindent\textbf{Grow Space $\bm\Omega$.} We use 4 stages for progressive learning. The initial scaling ratio $s_1$ is set to ${0.5}$ or $0.4$; the corresponding grow spaces are denoted by $0.5\Omega$ and $0.4\Omega$. By default, we use $0.5\Omega$ for our experiments, unless mentioned otherwise. The grow space of $n$ and $l$ are calculated by multiplying the value of the whole model with 4 equispaced scaling ratios $s \in \{0.5,0.67,0.83,1.0\}$, and we round the results to valid integer values. We use \textit{Prog} to denote our manual progressive baseline with \textit{uniform linear growth schedule} as described in Sec. \ref{sec:GS}.
\vspace{-0.6em}
\subsection{Efficient Training on ImageNet}\label{sec:exp_imagenet}
\vspace{-0.3em}
We first validate the effectiveness of AutoProg on ImageNet. As shown in \cref{tab:imagenet}, AutoProg consistently achieves remarkable efficient training results on diverse ViT architectures and training schedules.

\textbf{First}, our AutoProg achieves significant training acceleration over the regular training scheme with no performance drop. Generally, AutoProg speeds up ViTs training by more than 45\% despite changes on training epochs and network architectures. In particular, VOLO-D1 trained with \textit{AutoProg} 0.4$\Omega$ achieves \textbf{85.1\%} training acceleration, and even slightly improves the accuracy (+0.1\%).
\textbf{Second}, AutoProg outperforms the manual baseline, the uniform linear growing (Prog), by a large margin. For instance, Prog scheme causes severe performance degradation on DeiT-S. AutoProg improves over Prog scheme on DeiT-S by \textbf{1.7\%} on accuracy, successfully eliminating the performance gap by automatically choosing the proper growth schedule.
\textbf{Third}, as progressive learning uses smaller input size during training, one may question its generalization capability on larger input sizes. We answer this by directly testing the models trained with AutoProg on 288$\times$288 input size. The results justify that models trained with AutoProg have comparable generalization ability on larger input sizes to original models. Remarkably, VOLO-D1 trained for 300 epochs with AutoProg reaches \textbf{84.6\%} Top-1 accuracy when testing on 288$\times$288 input size, with \textbf{48.9\%} faster training.

The learning curves (\ie, evaluation accuracy during training) of DeiT-S and VOLO-D1 with different training schemes are shown in \cref{fig:learning_curve}.
Autoprog clearly accelerates the training progress of these two models. Interestingly, DeiT-S (100 epochs) trained with manual Prog scheme presents \textit{sharp fluctuations} after growth, while AutoProg successfully addresses this issue and eventually reaches higher accuracy by choosing proper growth schedule.
\vspace{-0.3em}
\subsection{Transfer Learning}
\vspace{-0.3em}
\begin{table}[t]
\setlength{\tabcolsep}{10pt}
\footnotesize
    \centering
    \begin{tabular}{lccc}
    \toprule
    Pretrain   & Speedup & CIFAR-10  & CIFAR-100\\
    \midrule
    Original    &\na        &99.0   &89.5  \\
    \rowcolor{Light}AutoProg    &\bf{48.9}\%     &\bf{99.0}   &\bf{89.7}   \\
    \bottomrule
    \end{tabular}
    \caption{Transfer learning results of DeiT-S on CIFAR datasets. The evaluation metric is Top-1 accuracy (\%).}
    \label{tab:transfer}
    \vspace{2pt}
\end{table}
To further evaluate the transfer ability of ViTs trained with AutoProg, we conduct transfer learning on CIFAR-10 and CIFAR-100 datasets.
We use the DeiT-S model that is pretrained with AutoProg on ImageNet for finetuning on CIFAR datasets, following the procedure in~\cite{touvron2020deit}. We compare with its counterpart pretrained with the ordinary training scheme.
The results are summarized in \cref{tab:transfer}. While AutoProg largely saves training time, it achieves competitive transfer learning results. This proves that AutoProg acceleration on ImageNet pretraining does not harm the transfer ability of ViTs on CIFAR datasets. 

\vspace{-0.3em}
\subsection{Ablation Study}\label{sec:exp_ablation}
\vspace{-0.3em}
\noindent\textbf{Growth Operator $\bm\zeta$.}
We first compare the three growth operators mentioned in \cref{sec:GO}, \ie, \textit{RandInit}~\cite{Simonyan2015VeryDC}, \textit{Stacking}~\cite{Gong2019EfficientTO} and \textit{Interpolation}~\cite{chang2018multi,Dong2020TowardsAR}, by using them with manual Prog scheme on VOLO-D1. As shown in \cref{tab:ablation_growth}, \textit{Interpolation} growth achieves the best accuracy both after the first growth and in the final. 

Then, we compare two growth operators build upon \textit{Interpolation} scheme, our proposed MoGrow, and Identity, which is a function-preserving~\cite{Chen2016Net2NetAL,Wei2016NetworkM} operator that can be achieved by Interpolation + ReZero~\cite{Bachlechner2020ReZeroIA}.
Specifically, ReZero uses a zero-initialized, learnable scalar to scale the residual modules in networks.
Using this technique on newly added layers can assure the original network function is preserved. The results are shown in \cref{tab:ablation_growth}. Contrary to expectations, we observe that Identity growth largely \textit{reduces} the Top-1 accuracy of VOLO-D1 (-3.21\%), probably because the network convergence is slowed down by the small scalar; besides, the global minimum of the original function could be a local minimum in the new network, which hinders the optimization. On this inferior growth schedule, our MoGrow still improves over Interpolation by 0.15\%, effectively reducing its performance gap.

Previous comparisons are based on the Prog scheme. Moreover, we also analyze the effect of MoGrow on AutoProg. The results are shown in \cref{tab:ablation_MoGrow}. We observe that MoGrow largely improves the performance of the supernet by \textbf{2.73\%}. It also increases the final training accuracy by 0.2\%, proving the effectiveness of MoGrow in AutoProg.

\begin{table}[t]
\footnotesize
    \centering
    \begin{tabular}{l|c|c}
    \toprule
    Growth Op. $\bm\zeta$                           & Top-1@Growth (\%)   & Top-1 (\%)  \\
    \midrule
    Baseline                                        &\na            & 82.53 \\
    \midrule
    RandInit~\cite{Simonyan2015VeryDC}              & 60.61         & 80.02 \\
    Stacking~\cite{Gong2019EfficientTO}              & 61.50         & 81.55 \\
    \rowcolor{Light}Interpolation~\cite{chang2018multi,Dong2020TowardsAR}& \textbf{61.53}& \textbf{81.78} \\
    \midrule
    Identity~\cite{Chen2016Net2NetAL,Wei2016NetworkM} & 61.04         & 79.32 \\
    \rowcolor{Light}MoGrow                          &\textbf{61.65} & \textbf{81.90} \\

    \bottomrule
    \end{tabular}%
    \vspace{-5pt}
    \caption{Ablation analysis of depth growth operator $\bm\zeta$ with the Prog learning scheme. Top-1@Growth denotes the accuracy after training for the first epoch of the second stage.}
    \label{tab:ablation_growth}
    \vspace{-5pt}
\end{table}

\begin{table}[t]
\setlength{\tabcolsep}{6pt}
\footnotesize
    \centering
    \begin{tabular}{l|c|c}
    \toprule
    Method   & Top-1@Growth (\%)  & Top-1 (\%)  \\
    \midrule
    AutoProg w/o MoGrow                             & 59.41                 & 82.6 \\
    \rowcolor{Light}AutoProg w/ MoGrow                              & \textbf{62.14}        & \textbf{82.8}\\
    \bottomrule
    \end{tabular}%
    \vspace{-5pt}
    \caption{Ablation analysis of \textit{MoGrow} in our AutoProg learning scheme on VOLO-D1. Top-1@Growth denotes the accuracy of the supernet after training for the first epoch of the second stage.}
    \label{tab:ablation_MoGrow}
    \vspace{5pt}
\end{table}

\noindent\textbf{Weight Recycling.} We further study the effect of weight recycling by training VOLO-D1 using AutoProg. As shown in \cref{tab:ablation_weightRec}, by recycling the weights of the supernet, AutoProg can achieve 12.3\% more speedup. Also, benefiting from the synergy effect in weight-nesting~\cite{yu2019slimmable}, weight recycling scheme does not cause accuracy drop. These results prove the effectiveness of weight recycling.

\noindent\textbf{Adaptive Regularization.}
Adaptive Regularization (AdaReg) for progressive learning is proposed in \cite{Tan2021EfficientNetV2SM}. It adaptively change regularization intensity (including RandAug~\cite{Cubuk2020RandaugmentPA}, Mixup~\cite{Zhang2018mixupBE} and Dropout~\cite{Srivastava2014DropoutAS}) according to network capacity of CNNs. Here, we generalize this scheme to ViTs and study its effect on ViT AutoProg training with DeiT-S and VOLO-D1. We mainly focus on three data augmentation and regularization techniques that are commonly used by ViTs, \ie, RandAug~\cite{Cubuk2020RandaugmentPA}, stochastic depth~\cite{Huang2016DeepNW} and random erase~\cite{Zhong2020RandomED}. When using AdaReg scheme, we linearly increase the magnitude of RandAug from 0.5$\times$ to 1$\times$ of its original value, and also linearly increase the probabilities of stochastic depth and random erase from 0 to their original values. The results of AutoProg with and without AdaReg are shown in \cref{tab:ablation_AdaReg}. Notably, DeiT-S can not converge when training with AdaReg, probably because DeiT models are heavily dependent on strong augmentations. \textit{On the contrary}, AdaReg on VOLO-D1 is \textit{indispensable}. Not using AdaReg causes 1.2\% accuracy drop on VOLO-D1. This result is consistent with previous discoveries on CNNs~\cite{Tan2021EfficientNetV2SM}. By default, we use AdaReg on VOLO models and not use it on DeiT models.

\begin{table}[t]
\setlength{\tabcolsep}{15pt}
\footnotesize
    \centering
    \begin{tabular}{l|c|c}
    \toprule
    Method  & Speedup   & Top-1 Acc. (\%)  \\
    \midrule
    w/o recycling   & 53.3\%       &   82.8    \\
    \rowcolor{Light}w/ recycling    &\textbf{65.6\%}    & \textbf{82.8}\\
    \bottomrule
    \end{tabular}
    \vspace{-5pt}
    \caption{Ablation analysis of \textit{weight recycling} in our AutoProg learning scheme on VOLO-D1.}
    \label{tab:ablation_weightRec}
    \vspace{-5pt}
\end{table}

\begin{table}[t]
\setlength{\tabcolsep}{5pt}
\footnotesize
    \centering
    \begin{tabular}{l|c|c|c}
    \toprule
    Method              & AdaReg    & Speedup   & Top-1 Acc. (\%)  \\
    \midrule
    \rowcolor{Light}DeiT-S AutoProg     & \xmark    &  \textbf{+40.7\%}  & \textbf{74.4}   \\
    DeiT-S AutoProg     & \cmark    &   \na     & \pzo0.1$^*$    \\
    \midrule
    VOLO-D1 AutoProg    & \xmark    &  +50.9\%  & 81.5  \\
    \rowcolor{Light}VOLO-D1 AutoProg    & \cmark    &  \textbf{+85.1\%}  & \textbf{82.7}  \\

    \bottomrule
    \end{tabular}
    \vspace{-2mm}
    \caption{Ablation analysis of the adaptive regularization on ViTs with the AutoProg learning scheme. (*: training can not converge)}
    \label{tab:ablation_AdaReg}
    \vspace{2pt}
\end{table}

\vspace{-5pt}
\section{Conclusion and Discussion}
\vspace{-5pt}
In this paper, we take a practical step towards sustainable deep learning by generalizing and automating progressive learning for ViTs. We have developed a strong manual baseline for progressive learning of ViTs with MoGrow growth operator and proposed an automated progressive learning (AutoProg) scheme for automated growth schedule search. Our AutoProg has achieved consistent training speedup on different ViT models with lossless performance on ImageNet and transfer learning. Ablation studies have proved the effectiveness of each component of AutoProg.

\noindent\textbf{Social Impact and Limitations.}
When network training becomes more efficient, it is also more available and less subject to regularization and study, which may result in a proliferation of models with harmful biases or intended uses.
In this work, we achieve inspiring results with automated progressive learning on ViTs. However, large scale training of CNNs and language models can not directly benefit from it. We encourage future works to develop automated progressive learning for efficient training in broader applications.

\vspace{-5pt}
\section*{Acknowledgement}
\vspace{-5pt}
This work was supported in part by Australian Research Council (ARC) Discovery Early Career Researcher Award (DECRA) under DE190100626 and National Key R\&D Program of China under Grant No. 2020AAA0109700.

{\small
\bibliographystyle{ieee_fullname}
\bibliography{egbib}
}

\renewcommand\thesection{\Alph{section}}
\renewcommand\thefigure{\Alph{figure}}
\renewcommand{\thetable}{\Roman{table}}

\clearpage
\emptythanks
\setcounter{footnote}{0}
\setcounter{figure}{0}
\setcounter{table}{0}

\appendix

\def\bx{\bm{x}}
\definecolor{seagreen}{RGB}{62,187,163}

\makeatletter
\def\@fnsymbol#1{\ensuremath{\ifcase#1\or \dagger\or \ddagger\or
   \mathsection\or \mathparagraph\or \|\or **\or \dagger\dagger
   \or \ddagger\ddagger \else\@ctrerr\fi}}
\makeatother

\title{Automated Progressive Learning for Efficient Training of Vision Transformers\\[7pt]\large{Supplementary Material}}
\author{%
Changlin Li\textsuperscript{1,2,3} \quad
Bohan Zhuang\textsuperscript{3}\thanks{Corresponding author.} \quad
Guangrun Wang\textsuperscript{4} \quad
Xiaodan Liang\textsuperscript{5} \quad
Xiaojun Chang\textsuperscript{2} \quad
Yi Yang\textsuperscript{6}\\
{\normalsize%
\textsuperscript{1}Baidu Research \quad
\textsuperscript{2}ReLER, AAII, University of Technology Sydney}\\
{\normalsize%
\textsuperscript{3}Monash University \quad
\textsuperscript{4}University of Oxford \quad
\textsuperscript{5}Sun Yat-sen University \quad
\textsuperscript{6}Zhejiang University}\\
{\tt\small%
changlinli.ai@gmail.com, bohan.zhuang@monash.edu, wanggrun@gmail.com,}\\
{\tt\small%
xdliang328@gmail.com, xiaojun.chang@uts.edu.au, yangyics@zju.edu.cn}
}
\maketitle

\appendix
\setlength{\textfloatsep}{2pt}

\section{Definition of Compared Growth Operators}
Given a smaller network $\bm\psi_s$ and a larger network $\bm\psi_\ell$, a growth operator $\bm\zeta$ maps the parameters of the smaller one $\bm\omega_s$ to the parameters of the larger one $\bm\omega_\ell$ by: $\bm\omega_\ell = \bm\zeta(\bm\omega_s)$. Let $\bm\omega_\ell^i$ denotes the parameters of the $i$-th layer in $\bm\psi_\ell$\footnote{In our default setting, $i$ begins from the layer near the classifier.}. We consider several $\bm\zeta$ in depth dimension that maps $\bm\omega_s$ to layer $i$ of $\bm\psi_\ell$ by: $\bm\omega_\ell^i = \bm\zeta(\bm\omega_s, i)$. 

\noindent\textbf{RandInit.} \textit{RandInit} copies the original layers in $\bm\psi_s$ and random initialize the newly added layers:
\begin{equation}
    \bm\zeta_\textit{RandInit}(\bm\omega_s, i) = \left\{
    \begin{aligned}
    \bm\omega_s^i, ~~~~~~~~~~~& i\leq l_s\\
    \textit{RandInit}, ~~& i > l_s.\\
    \end{aligned}
    \right.
\end{equation}

\noindent\textbf{Stacking.} \textit{Stacking} duplicates the original layers and directly stacks the duplicated ones on top of them:
\begin{equation}
    \bm\zeta_\textit{Stacking}(\bm\omega_s, i) = 
    \bm\omega_s^{i\bmod{l_s}}.
\end{equation}

\noindent\textbf{Interpolation.} \textit{Interpolation} interpolates new layers of $\bm\psi_\ell$ in between original ones and copy the weights from their nearest neighbor in $\bm\psi_s$:
\begin{equation}
    \bm\zeta_\textit{Interpolation}(\bm\omega_s, i) = \bm\omega_s^{\lfloor i/l_s\rfloor}.
\end{equation}

\section{Implementation Details}
Our ImageNet training settings follow closely to the original training settings of DeiT~\cite{touvron2020deit} and VOLO~\cite{Yuan2021VOLOVO}, respectively. We use the AdamW optimizer~\cite{loshchilov2018decoupled} with an initial learning rate of 1e-3, a total batch size of 1024 and a weight decay rate of 5e-2 for both architectures. The learning rate decays following a cosine schedule with 20 epochs warm-up for VOLO models and 5 epochs warm-up for DeiT models. For both architectures, we use exponential moving average with best momentum factor in $\{0.998, 0.9986, 0.999, 0.9996\}$.

For DeiT training, we use RandAugment~\cite{Cubuk2020RandaugmentPA} with 9 magnitude and 0.5 magnitude std., mixup~\cite{Zhang2018mixupBE} with 0.8 probability, cutmix~\cite{Yun2019CutMixRS} with 1.0 probability, random erasing~\cite{Zhong2020RandomED} with 0.25 probability, stochastic depth~\cite{Huang2016DeepNW} with 0.1 probability and repeated augmentation~\cite{Hoffer2020AugmentYB}.

For VOLO training, we use RandAugment~\cite{Cubuk2020RandaugmentPA}, random erasing~\cite{Zhong2020RandomED}, stochastic depth~\cite{Huang2016DeepNW}, token labeling with MixToken~\cite{jiang2021all}, with magnitude of RandAugment, probability of random erasing and stochastic depth adjusted by Adaptive Regularization.

\noindent\textbf{Adaptive Regularization.}
The detailed settings of Adaptive Regularization for VOLO progressive training is shown in \cref{tab:AdaReg}. These hyper-parameters are set heuristically regarding the model size. They perform fairly well in our experiments, but could still be sub-optimal.
\begin{table}[ht]
    \centering
    \footnotesize
    \setlength{\tabcolsep}{10pt}
    \vspace{-5pt}
    \begin{tabular}{l|cc|cc}
    \toprule
         \multirow{2}{*}{Regularization}&  \multicolumn{2}{c|}{D0} &   \multicolumn{2}{c}{D1}\\
         & min & max & min & max\\
         \midrule
         RandAugment~\cite{Cubuk2020RandaugmentPA} & 4.5 & 9 & 4.5 & 9\\
         Random Erasing~\cite{Zhong2020RandomED}& 0 & 0.25 & 0.0625 & 0.25\\
         Stoch. Depth~\cite{Huang2016DeepNW} & 0 & 0.1 & 0.1 & 0.2\\
    \bottomrule
    \end{tabular}\vspace{-5pt}
    \caption{Adaptive Regularization Settings (magnitude of RandAugment~\cite{Cubuk2020RandaugmentPA}, probability of Random Erasing~\cite{Zhong2020RandomED} and Stochastic Depth~\cite{Huang2016DeepNW}) for progressive training of VOLO models.}
    \label{tab:AdaReg}
    \vspace{-5pt}
\end{table}

\begin{table*}[t]
    \small
    \centering
    \setlength{\tabcolsep}{7pt}
        \begin{tabular}{l|l|cc|cc|c|c}
        \toprule
        Model  & \makecell[l]{Training\\scheme}  &\makecell{FLOPs\\(avg. per step)} & Speedup & \makecell{Runtime\\(GPU Hours)} & Speedup    & \makecell{Top-1\\(\%)} & \makecell{Top-1@288\\(\%)}\\
        \midrule
        \multicolumn{4}{l}{\textbf{\textit{100 epochs}}}\\
        \midrule
        \multirow{3}{*}{DeiT-S~\cite{touvron2020deit}}          & Original     &   4.6G & \na &71& \na           & 74.1      & 74.6\\ %
        & Prog       &2.4G&+91.6\%&                            46
        & +53.6\% & 72.6      &  73.2\\ %
        & \CC AutoProg                                             &\CC 2.8G&\CC +62.0\%&\CC  50  &\CC +40.7\%                   & \CC \textbf{74.4}      &\CC \textbf{74.9}\\ %
        \arrayrulecolor{lightgray}\hline\arrayrulecolor{black}
        \multirow{4}{*}{VOLO-D1~\cite{Yuan2021VOLOVO}}          & Original     &6.8G&\na&150& \na           & 82.6      &83.0\\ %
        & Prog                                                     &3.7G&+84.7\%&93& +60.9\% & 81.7      &82.1\\ %
        & \CC AutoProg 0.5$\Omega$                                     &\CC 3.3G&\CC +104.2\%&\CC 91&\CC +65.6\%                   &\CC \textbf{82.8}      &\CC \textbf{83.2}\\ %
        & \CC AutoProg 0.4$\Omega$                                     &\CC 2.9G&\CC \textbf{+132.2\%}&\CC 81& \CC \textbf{+85.1\%}          &\CC 82.7      &\CC 83.1 \\ %
        \arrayrulecolor{lightgray}\hline\arrayrulecolor{black}
        \multirow{3}{*}{VOLO-D2~\cite{Yuan2021VOLOVO}}          & Original     &14.1G&\na&277& \na                       & 83.6      & 84.1\\ %
        & Prog                                                     &7.5G&+87.7\%&180& +54.4\% & 82.9      & 83.3\\ %
        &\CC AutoProg                                                 &\CC 8.3G&\CC +68.7\%&\CC191&\CC +45.3\%                   &\CC \textbf{83.8}      &\CC \textbf{84.2}\\ %
        \midrule
        \multicolumn{4}{l}{\textbf{\textit{300 epochs}}}\\
        \midrule
        \multirow{2}{*}{DeiT-Tiny~\cite{touvron2020deit}}       &  Original    &1.2G&\na&144& \na           & 72.2      & 72.9\\ %
        & \CC  AutoProg                                                 &\CC 0.7G&\CC +82.1\%&\CC 95&\CC     +51.2\%                       &\CC   \textbf{72.4}  &\CC  \textbf{73.0}  \\ %
        \arrayrulecolor{lightgray}\hline\arrayrulecolor{black}
        \multirow{2}{*}{DeiT-S~\cite{touvron2020deit}}          &  Original    &4.6G&\na&213& \na           & 79.8      &  80.1\\ %
        & \CC AutoProg                                                 &\CC 2.8G&\CC +62.0\%&\CC 150&\CC  +42.0\%                   & \CC 79.8      & \CC  80.1 \\ %
        \arrayrulecolor{lightgray}\hline\arrayrulecolor{black}
        \multirow{2}{*}{VOLO-D1~\cite{Yuan2021VOLOVO}}          & Original     &6.8G&\na&487& \na           & 84.2      & 84.4 \\ %
        & \CC AutoProg                                                 &\CC 4.0G&\CC +68.9\%&\CC 327&\CC  +48.9\%                   &\CC  \textbf{84.3}      &\CC \textbf{84.6}\\ %
        \arrayrulecolor{lightgray}\hline\arrayrulecolor{black}
        \multirow{2}{*}{VOLO-D2~\cite{Yuan2021VOLOVO}}          &Original      &14.1G&\na&863& \na           & 85.2      & 85.1\\
        &\CC AutoProg                                                 &\CC 8.8G&\CC +60.7\%&\CC 605&\CC  +42.7\%                   &\CC  85.2     & \CC  \textbf{85.2}\\
        \bottomrule
    \end{tabular}%
    \caption{Detailed results of efficient training on ImageNet. Best results are marked with \textbf{Bold}; our method or default settings are highlighted in \colorbox{Light}{purple}. Top-1@288 denotes Top-1 Accuracy when directly testing on 288$\times$288 input size, \textit{without} finetuning. Runtime is rounded to integer.}
    \label{tab:appimagenet}
    \vspace{-10pt}
\end{table*}

\noindent\textbf{Growth Space $\bm\Lambda_{\bm{k}}$ in Each Stage.}
We find emprically that the elastic supernet converges faster when the number of sub-networks are smaller. Thus, restricting the growth space $\Lambda_k$ in each stage could help the convergence of the supernet. In practice, we make the restriction that $|\Lambda_k|\leq 9$. Specifically,
in the first stage, we use the largest, the smallest and the medium candidates of $n$ and $l$ in $\Omega$ to construct $\Lambda_1$, which makes it possible to route to the whole network and perform regular training if the growing ``ticket'' (suitable sub-network) does not exist. In each of the following stages, we include the next 3 candidates of $l$ and the next 1 candidate of $n$, forming a growth space with $2\times 4 = 8$ candidates.

\section{Additional Results}
\vspace{-5pt}
\noindent\textbf{Theoretical Speedup.} In \cref{tab:appimagenet}, we calculate the average FLOPs per step of different learning schemes. AutoProg consistently achieves more than 60\% speedup on theoretical computation. Remarkably, VOLO-D1 trained for 100 epochs with AutoProg 0.4$\Omega$ achieves \textbf{132.2\%} theoretical acceleration. The gap between theoretical and practical speedup indicates large potential of AutoProg. We leave the further improvement of practical speedup to future works; for example, AutoProg can be further accelerated by adjusting the batch size to fill up the GPU memory during progressive learning.

\noindent\textbf{Comparison with Progressively Stacking.}
Progressively Stacking~\cite{Gong2019EfficientTO} (ProgStack) is a popular progressive learning method in NLP to accelerate BERT pretraining. It begins from $\frac{1}{4}$ of original layers, then copies and stacks the layers twice during training. Originally, it has three training stages with number of steps following a ratio 5:7:28. In CompoundGrow~\cite{Gu2021OnTT}, this baseline is implemented as three stages with 3:4:3 step ratio. Our implementation follows closer to the original paper, using a ratio of 1:2:5.
The results are shown in \cref{tab:progstack}. ProgStack achieves relatively small speedup with performance drop (0.4\%). Our MoGrow reduces this performance gap to 0.1\%. AutoProg achieves 74.1\% more speedup and 0.5\% accuracy improvement over the ProgStack baseline.

\begin{table}[t]
    \centering
    \small
    \setlength{\tabcolsep}{4pt}
    \begin{tabular}{l|cc|c}
    \toprule
    Training scheme  & \makecell{Runtime\\(GPU hours)}& Speedup & Top-1 (\%)\\
    \midrule
    Baseline     & ~150.2~           & -& 82.6 \\
    \arrayrulecolor{lightgray}\hline\arrayrulecolor{black}
    ProgStack~\cite{Gong2019EfficientTO} &135.3 & +11.0\% & 82.2 \\
     + MoGrow & 136.0 & +10.4\% & 82.5 \\
     \arrayrulecolor{lightgray}\hline\arrayrulecolor{black}
     Prog & 93.3 & +60.9\% & 81.7\\
     \rowcolor{Light}AutoProg 0.4$\Omega$  & 81.1     & \textbf{+85.1\%}& \textbf{82.7}\\
     \bottomrule
    \end{tabular}
    \caption{Comparison with progressively stacking.}\vspace{5pt}
    \label{tab:progstack}
\end{table}

\noindent\textbf{Combine with AMP.}
Automatic mixed precision (AMP) [\textcolor{green}{52}] is a successful and mature low-bit precision efficient training method.
We conduct experiments to prove that the speed-up achieved by AutoProg is orthogonal to that of AMP. As shown in \cref{tab:amp}, the relative speed-up achieved by AutoProg with or without AMP is comparable (+85.1\% \vs +87.5\%), proving the orthogonal speed-up.
\begin{table}[ht]
    \centering
    \small
    \setlength{\tabcolsep}{4pt}
    \begin{tabular}{l|c|c}
    \toprule
    Method          & Speed-up  & Top-1 Acc. (\%)  \\
    \midrule
    Original (w/o AMP)& \na       & 82.6 \\
    AMP             & +74.0\%   & 82.6 \\
    AutoProg        &\cellcolor{Light} \textbf{+87.5\%}   & \textbf{82.7} \\
    \midrule
    \multirow{2}{*}{AMP + AutoProg}  & \textbf{+222.1\%}   & \multirow{2}{*}{\textbf{82.7}}\\
    &\cellcolor{Light}(\textbf{+85.1\%} over AMP)&\\
    \bottomrule
    \end{tabular}
    \vspace{-5pt}\caption{Speed-up of AutoProg is orthogonal to AMP [\textcolor{green}{52}].}
    \label{tab:amp}
\end{table}

\noindent\textbf{Number of stages.}
We perform experiments to analyze the impact of the number of stages on AutoProg with different initial scaling ratios (0.5 and 0.4). As shown in \cref{tab:stage}, AutoProg is not very sensitive to stage number settings. Fewer than 4 yields more speed-up, but could damage the performance. In general, the default 4 stages setting performs the best. When scaling the stage number to 50, there are only supernet training phases (2 epochs per stage) during the whole 100 epochs training, causing severe performance degradation.

\begin{table}[ht]
    \centering
    \scriptsize
    \setlength{\tabcolsep}{4pt}
    \begin{tabular}{c|l|c|c|c|c|c}
    \toprule
    Ratio & Num. Stages        & Orig.  & 3        & \cellcolor{Light}4        & 5             & 50  \\
    \midrule
    \multirow{2}{*}{0.5}&Speed-up        & \na    & \textbf{+69.1\%}   & \cellcolor{Light}+65.6\%  & +63.6\%       & +48.5\% \\
    &Top-1 Acc. (\%) & 82.6   & 82.6     & \cellcolor{Light}\textbf{82.8}     & \textbf{82.8}          & 81.7 \\
    \midrule
    \multirow{2}{*}{0.4}&Speed-up        & \na    & \textcolor{gray}{+90.8\%}   & \cellcolor{Light}\textbf{+85.1\%}  & +80.4\%       & \na \\
    &Top-1 Acc. (\%) & 82.6   & 82.4     & \cellcolor{Light}\textbf{82.7}     & \textbf{82.7}          & \na \\
    \bottomrule
    \end{tabular}
    \caption{Ablation analysis on number of stages.}
    \label{tab:stage}
\end{table}

\noindent\textbf{Effect of Progressive Learning in AutoProg.} AutoProg is comprised by its two main components, ``Auto'' and ``Prog''. The effectiveness of ``Auto'' is already studied by comparing with Prog in the main text. Here, we study the effectiveness of progressive learning in AutoProg by training an elastic supernet baseline for 100 epochs without progressive growing to compare with AutoProg. Specifically, we treat VOLO-D1 as an Elastic Supernet, and train it by randomly sampling one of its sub-networks in each step, same to the search stage in AutoProg. The results are shown in \cref{tab:ablation_prog}. In previous works that uses elastic supernet~\cite{yu2019slimmable,Yu2020BigNASSU,chen2021autoformer}, the supernet usually requires more training iterations to reach a comparable performance to a single model. As expected, the supernet performance is lower than the original network given the same training epochs. Specifically, AutoProg improves over elastic supernet baseline by 1.1\% Top-1 accuracy, with 17.1\% higher training speedup, reaching the performance of the original model with the same training epochs but much faster, which proves the superiority of progressive learning. 
\begin{table}[ht]
\setlength{\tabcolsep}{15pt}
\footnotesize
    \centering
    \begin{tabular}{l|c|c}
    \toprule
    Method      & Speedup           & Top-1 Acc. (\%)  \\
    \midrule
    Original    & \na               &   82.6    \\
    Supernet    & 48.5\%           &   81.7    \\
    \rowcolor{Light}AutoProg    &\textbf{65.6\%}    & \textbf{82.8}\\
    \bottomrule
    \end{tabular}
    \caption{Ablation analysis of progressive learning in AutoProg with VOLO-D1.}
    \label{tab:ablation_prog}
\end{table}

\noindent\textbf{Analyse of Searched Growth Schedule.}
Two typical growth schedules searched by AutoProg are shown in \cref{tab:searched_schedule}. AutoProg clearly prefers smaller token number than smaller layer number. Nevertheless, selecting a small layer number in the first stage is still a good choice, as both of the two schemes use reduced layers in the first stage.

\begin{table}[ht]
    \centering
    \footnotesize
    \begin{tabular}{l|c|c|c|c|c}
        \toprule
         \multicolumn{2}{c|}{Stage $k$}  & 1     &2      &3      &4  \\
         \midrule
         \multirow{2}{*}{\makecell{VOLO-D1 100e 0.4$\Omega$}}&$l$        & 0.4   &1      &1      &1  \\
         \arrayrulecolor{lightgray}\cline{2-6}\arrayrulecolor{black}
         &$n$        & 0.4   &0.6    &0.6    &1  \\
        \hline
        \multirow{2}{*}{\makecell{VOLO-D2 300e}}&$l$        & 0.83  &1      &1      &1  \\
        \arrayrulecolor{lightgray}\cline{2-6}\arrayrulecolor{black}
         &$n$        & 0.5   &0.67   &0.83   &1  \\
         \bottomrule
    \end{tabular}
    \caption{Searched growth schedules for VOLO-D1 0.4$\Omega$, 100 epochs, and VOLO-D2, 300 epochs.}
    \label{tab:searched_schedule}
\end{table}

\noindent\textbf{Retraining with Searched Growth Schedule.}
To evaluate the searched growth schedule, we perform retraining from scratch with VOLO-D1, using the schedule searched by AutoProg 0.4$\Omega$. As shown in \cref{tab:retrain}, retraining takes slightly longer time (-0.6\% speedup) because the speed of searched optimal sub-networks could be slightly slower than the average speed of sub-networks in the elastic supernet. Retraining reaches the same final accuracy, proving that the searched growth schedule can be used separately.

\begin{table}[ht]
    \centering
    \footnotesize
    \setlength{\tabcolsep}{4pt}
    \begin{tabular}{l|cc|c}
    \toprule
    Training scheme     &\makecell{Runtime\\(GPU hours)} & Speedup & Top-1 (\%) \\
    \midrule
        Baseline    & ~150.2~           & -& 82.6 \\
    \arrayrulecolor{lightgray}\hline\arrayrulecolor{black}
    \rowcolor{Light}AutoProg 0.4$\Omega$ & 81.1     & \textbf{+85.1\%}& \textbf{82.7}\\
    Retrain & 81.4 & +84.5\% & \textbf{82.7}\\
    \bottomrule
    \end{tabular}
    \caption{Retraining results with searched growth schedule on VOLO-D1, 100 epochs.}
    \label{tab:retrain}
\end{table}

\noindent\textbf{Extend to CNNs.} To explore the effect of our policy on CNNs, we conduct experiments with ResNet50 \cite{he2016deep}, and found that the policy searched on ViTs generalizes very well on CNNs (see \cref{tab:r50}). These results imply that AutoProg opens an interesting direction (automated progressive learning) to develop more general learning methods for a wide computer vision field.

\begin{table}[ht]
    \centering
    \scriptsize
    \setlength{\tabcolsep}{18pt}
    \vspace{-5pt}
    \begin{tabular}{l|c|c}
    \toprule
    Method          & Speed-up  & Top-1 Acc. (\%)  \\
    \midrule
    Original        & \na       & 77.3 \\
    \rowcolor{Light}AutoProg        & \textbf{+56.9\%}   & 77.3 \\
    \bottomrule
    \end{tabular}
    \caption{AutoProg with ResNet50 \cite{he2016deep} on ImageNet (100 epochs).}
    \vspace{-13pt}
    \label{tab:r50}
\end{table}
\vfill

\end{document}